\documentclass[conference]{IEEEtran}
\IEEEoverridecommandlockouts
\usepackage{cite}
\usepackage{amsmath,amssymb,amsfonts}
\usepackage{algorithmic}
\usepackage{graphicx}

\usepackage{pifont}
\usepackage{textcomp}
\usepackage{arydshln}
\usepackage{url}
\usepackage{xcolor}
\usepackage{subcaption}
\usepackage{multirow}
\def\BibTeX{{\rm B\kern-.05em{\sc i\kern-.025em b}\kern-.08em
    T\kern-.1667em\lower.7ex\hbox{E}\kern-.125emX}}
\begin{document}

\title{GlaBoost: A Multimodal Structured Framework for Glaucoma Risk Stratification\\
\thanks{Dr. Cheng Huang, Zeyu Han, Weizheng Xie and Dr. Jia Zhang are with the Department of Computer Science, Southern Methodist University, Dallas, TX 75205, USA.}

\thanks{Dr. Karanjit Kooner and Dr. Jui-Kai Wang are with the Department of Ophthalmology, UT Southwestern Medical Center, Dallas, TX 75390, USA.}

\thanks{Dr. Tsengdar Lee is with the High Performance Computing Program, National Aeronautics and Space Administration, Washington, DC 20546, USA.}

\thanks{$^{\dag}$Corresponding Author: jiazhang@smu.edu}
}
\author{
\IEEEauthorblockN{Cheng Huang, Zeyu Han, Weizheng Xie, Karanjit Kooner, Tsengdar Lee, Jui-Kai Wang, Jia Zhang\textsuperscript{$\dag$} }
}

\maketitle

\begin{abstract}
Early and accurate glaucoma detection is critical to prevent 
irreversible vision loss, yet existing AI methods often rely on 
unimodal inputs and lack interpretability. We present 
GlaBoost, a multimodal gradient boosting framework that 
unifies three complementary signals for glaucoma risk prediction: 
fundus image embeddings from a pretrained convolutional encoder, 
free-text neuroretinal rim assessments encoded by a 
transformer-based language model, and structured ophthalmic 
biomarkers. These modalities are fused into a single 
representation and classified by an enhanced XGBoost model. 
On two real-world annotated datasets, GlaBoost consistently outperforms 
unimodal and generic multimodal baselines. Feature importance analysis highlights the cup-to-disc ratio, rim thinning, and the ISNT rule as the dominant predictors,  yielding clinically consistent and interpretable decisions. GlaBoost offers a transparent and scalable foundation for 
multimodal decision support in ophthalmology.
\end{abstract}


\section{Introduction}

Glaucoma is a progressive optic neuropathy that ranks among the 
leading causes of irreversible blindness worldwide 
\cite{lag,lag2,bg-2}. Early and accurate diagnosis is critical 
but remains challenging due to the disease's complex presentation 
and its reliance on heterogeneous diagnostic 
evidence-fundus images \cite{octa500}, anatomical 
measurements \cite{glagan}, and expert clinical descriptions 
\cite{svaml}.

Recent AI-based approaches, particularly those built on 
convolutional neural networks (CNNs), have achieved strong 
performance in image-based glaucoma 
detection~\cite{fairseg,fairvision,glagan,unet,gdp,fairdomain}. 
However, these systems typically operate on a single visual 
modality and fail to incorporate the structured biomarkers and 
free-text annotations that ophthalmologists routinely 
rely on~\cite{aensi,svaml}. This unimodal bias limits both 
diagnostic accuracy and clinical interpretability.

To bridge this gap, we propose GlaBoost, a multimodal 
learning framework that fuses fundus image embeddings, structured 
ophthalmic biomarkers, and transformer-encoded clinical narratives 
within an enhanced XGBoost classifier~\cite{xgboost}. By 
explicitly modeling expert textual descriptions alongside visual 
and numeric features, GlaBoost captures the semantic context that 
underlies clinical reasoning.

Our main contributions are summarized as follows:
\begin{itemize}
  \item We propose the GlaBoost, a multimodal 
  XGBoost-based framework that jointly leverages fundus images, 
  structured anatomical biomarkers, and free-text clinical 
  descriptions for glaucoma risk prediction.
  \item We design an effective pipeline for encoding free-text 
  ophthalmic observations via a pretrained language model, and 
  quantify their contribution to diagnostic performance through 
  ablation studies.
  \item We demonstrate, across two real-world datasets and a 
  comprehensive set of unimodal, multimodal, and large language 
  model baselines, that multimodal fusion improves both 
  predictive accuracy and feature-level interpretability in 
  glaucoma diagnosis.
\end{itemize}

\section{Related Work}

\subsection{AI for Glaucoma}
Most existing glaucoma AI systems are built around 
CNN-based architectures~\cite{lag,lag2}, including U-Net 
variants~\cite{unet}, and have been applied to detection, 
segmentation, and reconstruction across diverse retinal imaging 
modalities~\cite{fairclip,fairvision,glagan,fairseg,octa500}. 
Recent progress in large language models (LLMs) and 
vision-language models (VLMs) has further enabled the 
incorporation of textual evidence into medical 
AI~\cite{gpt4o,llama3,r1,v3}. 
Nevertheless, current glaucoma AI remains predominantly 
image-driven~\cite{fairvision,glagan,fairseg}, while clinical 
diagnosis depends on interpretable biomarkers such as 
intraocular pressure (IOP) and cup-to-disc ratio~\cite{bg-1,bg-2}, a 
mismatch that constrains both clinical adoption and trust.

\subsection{Clinical Decision Making}
In practice, ophthalmologists synthesize fundus imaging, 
patient history, and free-text descriptions of optic nerve 
head morphology to reach a diagnosis~\cite{bg-1,bg-2}. 
Such textual descriptions encode semantic cues, e.g., rim 
pallor, notching, and bayoneting, that are difficult to 
recover from pixel-level features 
alone~\cite{aensi,svaml}. 
Although recent explainable AI techniques offer interpretability 
through feature attribution or attention-based 
visualization~\cite{att}, they are typically applied to a single 
modality and therefore cannot align with the inherently 
multimodal nature of clinical reasoning. 
GlaBoost is designed to close this gap by jointly modeling 
imaging, structured biomarkers, and expert narratives within a 
single interpretable framework.

\section{Methodology: GlaBoost}

\noindent\textbf{Data Input}: let $\mathcal{D} = \{(\mathbf{x}_i, y_i)\}_{i=1}^N$ denote the dataset, where each instance consists of multimodal features $\mathbf{x}_i$ and a binary label $y_i \in \{0,1\}$ indicating the presence of glaucoma. We decompose the input feature vector $\mathbf{x}_i$ as:
\begin{equation}
    \mathbf{x}_i = \left[\mathbf{x}_i^{\text{text}}; \mathbf{x}_i^{\text{struct}}; \mathbf{x}_i^{\text{human}}; \mathbf{x}_i^{\text{img}}\right]
    \label{eq1}
\end{equation}

where $\mathbf{x}_i^{\text{text}} \in \mathbb{R}^{d_t}$ is the textual embedding, $\mathbf{x}_i^{\text{struct}} \in \mathbb{R}^{d_s}$ represents structured clinical features, and $\mathbf{x}_i^{\text{img}} \in \mathbb{R}^{d_v}$ are visual features extracted from fundus images. $\mathbf{x}_i^{\text{human}}$ is the data based on human subjective judgment, not a label, and in certain datasets, it is not required and is considered optional.

\begin{figure}[ht]
    \centering
    \includegraphics[width=1.0\linewidth]{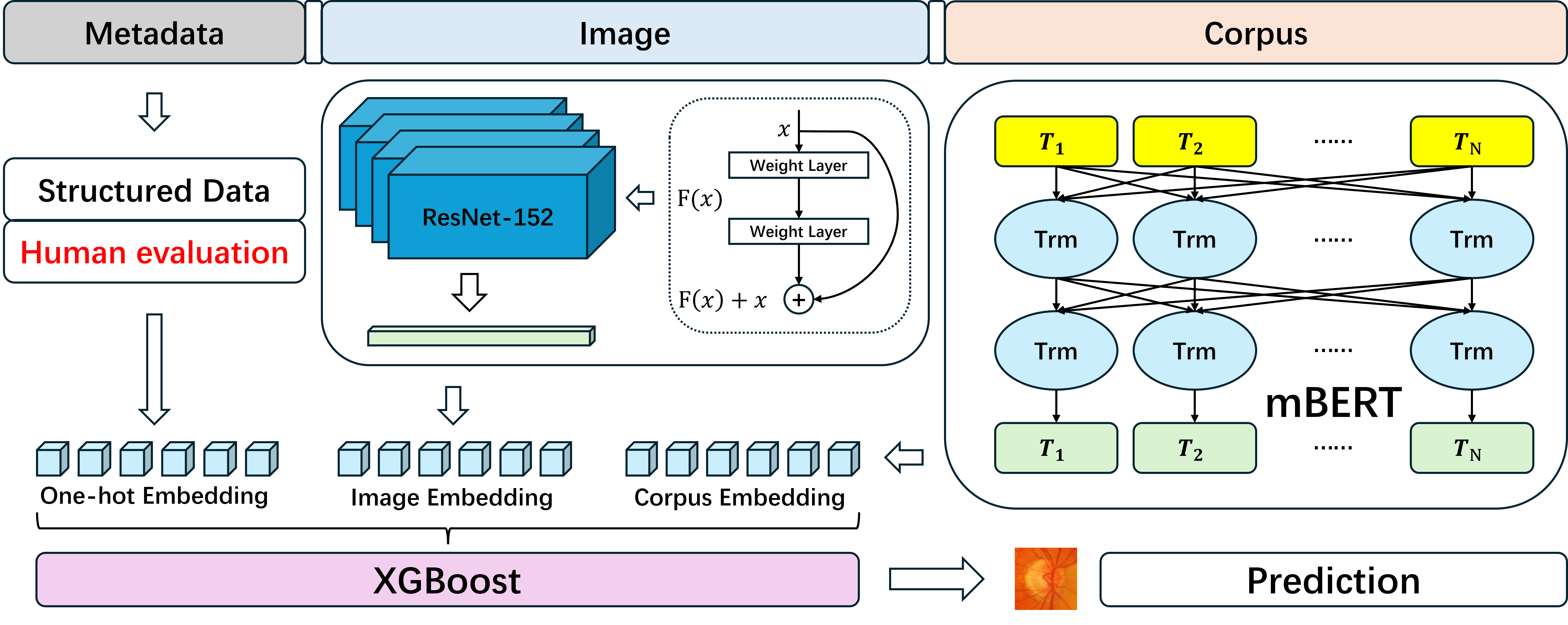}
    \caption{The Architecture of GlaBoost}
    \label{fig:placeholder}
\end{figure}

Based on this input of data, as shown in Fig.~\ref{fig:placeholder} GlaBoost have 4 stages: (1) \textbf{Structured Feature Extraction}; (2) \textbf{Image Feature Extraction}; (3) \textbf{Textual Feature Extraction} and (4) \textbf{Multimodal Fusion and Classification}.

\subsection{Structured Feature Extraction}
Each instance contains a clinical report field description where we extract two or one types of structured features:

\subsubsection{Machine-Readable Fundus Biomarker}
Within each report, the \texttt{fundus\_features} field provides structured annotations of retinal morphology. 
We extract clinically established indicators, including optic disc size (small, normal, large) and the cup-to-disc ratio ($\in[0,1]$). 
Categorical features are converted via one-hot encoding, and missing values are explicitly encoded as additional categories to ensure model compatibility, as formulated in Eq.~\ref{eq2}:

\begin{equation}
    \mathbf{x}_i^{\text{struct}} = \Phi \big( g(\text{description}_i) \big) \in \mathbb{R}^{d_s},
    \label{eq2}
\end{equation}

where $g(\cdot)$ extracts raw fundus-related attributes, and $\Phi(\cdot)$ denotes a preprocessing function that performs one-hot encoding, normalization of continuous values, and explicit handling of missing entries. 
The resulting vector $\mathbf{x}_i^{\text{struct}}$ provides a fixed-dimensional representation of structured clinical information.

\subsubsection{Human-Evaluated Indicator (Optional)}
We extract two subjective diagnostic indicators provided by clinical annotators: \texttt{Glaucoma\_risk\_assessment} and \texttt{Confidence\_level}. 
The glaucoma risk assessment is treated as a categorical judgment and encoded using one-hot vectors, while the confidence level is retained as a continuous score in $[0,1]$. 
Both indicators are processed using the same feature extraction formulation as the structured data stage to obtain $\mathbf{x}_i^{\text{human}}$.

\subsection{Image Feature Extraction}
We apply a pretrained CNN, ResNet-152 \cite{resnet} to extract image features from each retinal fundus photograph $I_i$:

\begin{equation}
    \mathbf{x}_i^{\text{img}} = f_{\text{img}}(I_i; \theta_{\text{resnet}}) \in \mathbb{R}^{d_v}
    \label{eq3}
\end{equation}
The network is initialized with ImageNet-pretrained weights and extracts high-level image representations from the penultimate layer. 
All convolutional layers are frozen to reduce overfitting, with only the final projection layers optionally fine-tuned. 
Input images are resized to $224\times224$, normalized using ImageNet statistics, and optionally augmented during training with random horizontal flips and color jitter.

\subsection{Textual Feature Extraction}
Each clinical report contains unstructured narrative fields such as some description, which often include nuanced spatial and qualitative assessments of optic disc morphology. To encode this textual modality, we use a Transformer-based language encoder $f_{\text{text}}$, mBERT \cite{bert}:

\begin{equation}
    \mathbf{x}_i^{\text{text}} = f_{\text{text}}(T_i) = \text{mean-pool} \left( \text{Transformer}(T_i) \right) \in \mathbb{R}^{d_t}
    \label{eq4}
\end{equation}
where $T_i$ denotes the input text token sequence and mean-pooling aggregates contextualized token embeddings. The model is initialized with domain-general weights and remains frozen to avoid overfitting due to limited textual supervision. We apply the following preprocessing to $T_i$: (1) Lowercasing and punctuation normalization; (2) Truncation or padding to a maximum sequence length and (3) Removal of medically irrelevant boilerplate text (optional).

\subsection{Multimodal Fusion and Classification}

The final feature vector is constructed by concatenating all modalities:
\begin{equation}
    \mathbf{z}_i = \left [\mathbf{x}_i^{\text{text}}; \mathbf{x}_i^{\text{struct}}; \mathbf{x}_i^{\text{human}}; \mathbf{x}_i^{\text{img}}\right] \in \mathbb{R}^{d}
    \label{eq5}
\end{equation}

We use an XGBoost classifier \cite{xgboost} $f_\theta$ to predict the label:
\begin{equation}
\hat{y}_i = f_\theta(\mathbf{z}_i)
\end{equation}
and optimize it with logistic loss:
\begin{equation}
\mathcal{L}(\theta) = -\sum_{i=1}^{N} \left[ y_i \log \hat{y}_i + (1 - y_i)\log(1 - \hat{y}_i) \right] + \Omega(f_\theta)
\end{equation}
where $\Omega(f_\theta)$ is a regularization term controlling model complexity.

\subsection{Dataset}

\noindent\textbf{Ethics and Data Governance.}
The experimental procedures involving human subjects described in 
this paper were approved by the Institutional Review Board (IRB) 
of the \emph{University of Texas Southwestern Medical Center} 
(UTSW), in accordance with the Helsinki Declaration of 1975 
(revised in 2000). As the study is retrospective, the IRB waived 
the requirement for written informed consent. All patient records 
and fundus images used in this work were fully de-identified prior 
to analysis, and no protected health information was accessible to 
the modeling pipeline.

\begin{figure}[ht]
    \centering
    \begin{subfigure}[t]{0.35\linewidth}
        \includegraphics[width=\linewidth]{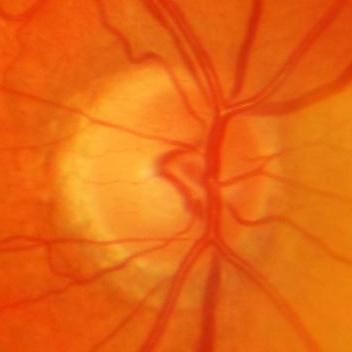}
        \caption{Public Dataset}
        \label{pd-1}
    \end{subfigure}
    \hfill
    \begin{subfigure}[t]{0.35\linewidth}
        \includegraphics[width=\linewidth]{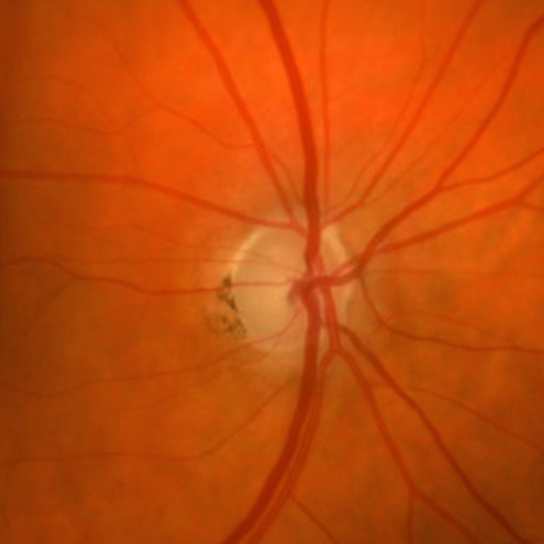}
        \caption{Private Dataset}
        \label{pd-2}
    \end{subfigure}
    \caption{The Sample of Image from Public and Private Dataset}
    \label{pd}
\end{figure}

\subsubsection{Public Dataset} 

TABLE~\ref{data-public} shows a representative sample from the 
public dataset, which is organized as JSON-formatted clinical 
reports paired with fundus images. During preprocessing, 
categorical fields such as \texttt{optic\_disc\_size} and 
continuous fields such as \texttt{cup\_to\_disc\_ratio} are mapped 
to binary indicators (\texttt{True}/\texttt{False}) following 
established clinical thresholds for glaucoma 
diagnosis~\cite{bg-1}. The free-text field 
\texttt{neuroretinal\_rim} is retained as an unstructured 
description and processed through the textual encoder.

The two fields \texttt{glaucoma\_risk\_assessment} and 
\texttt{confidence\_level} encode an annotator's 
\emph{preliminary risk tier} and \emph{self-rated confidence}, 
respectively. We treat them as auxiliary clinician-derived inputs 
rather than ground-truth labels. 
A representative fundus image is shown in Fig.~\ref{pd-1}. The 
dataset is publicly available\footnote{\url{https://huggingface.co/datasets/AswanthCManoj/glaucoma_diagnosis_json_analysis}}.

\begin{table}[h]
\caption{Representative Sample from the Public Dataset}
\begin{center}
\scalebox{0.95}{
\begin{tabular}{l|l}
\hline
\textbf{Field} & \textbf{Value} \\
\hline
\texttt{optic\_disc\_size}        & large \\
\texttt{cup\_to\_disc\_ratio}     & 0.8 \\
\texttt{isnt\_rule\_followed}     & false \\
\texttt{rim\_pallor}              & true \\
\texttt{rim\_color}               & pale \\
\texttt{bayoneting}               & true \\
\texttt{sharp\_edge}              & true \\
\texttt{laminar\_dot\_sign}       & true \\
\texttt{notching}                 & true \\
\texttt{rim\_thinning}            & true \\
\texttt{additional\_observations} & null \\
\texttt{neuroretinal\_rim}        & (free-text description) \\
\hline
\texttt{glaucoma\_risk\_assessment} & high risk \\
\texttt{confidence\_level}        & 0.9 \\
\hline
\end{tabular}}
\label{data-public}
\end{center}
\end{table}

\subsubsection{Private Dataset}
\label{sec:private_dataset}

TABLE~\ref{data-utsw} shows a representative sample from the 
private UTSW dataset, which is derived from clinical OCT reports. Each record contains 16 
biomarkers organized into three anatomical 
categories, RNFL, ONH, and GCC, measured 
independently for OD and OS eyes, together 
with the corresponding IE.

\noindent\textbf{Status Encoding} Each biomarker is associated 
with a clinical status assigned by the OCT analyzer: 
\emph{Within Normal}, \emph{Borderline}, or \emph{Outside Normal}. 
Borderline values lie within the normal reference range but 
approach the pathological boundary, and therefore carry 
diagnostic significance that is distinct from both ends. We 
encode each biomarker status as an ordinal score 
$\{0, 0.5, 1\}$ corresponding to \{Outside Normal, Borderline, 
Within Normal\}, preserving the underlying clinical ordering. 
Unless a biomarker is explicitly marked as Within Normal in the 
report, Borderline cases are retained as a distinct third 
category rather than being collapsed into the Normal class.

\noindent\textbf{Annotation} All UTSW records are reviewed by 
two board-certified glaucoma specialists from the Department of 
Ophthalmology, UT Southwestern Medical Center, who provide the 
final diagnostic labels used as ground truth in this study. A 
representative fundus image is shown in Fig.~\ref{pd-2}.

\begin{table}[h]
\caption{Sample from the UTSW Private Dataset}
\begin{center}
\scalebox{0.75}{
\begin{tabular}{l|l|c|c|c}
\hline
\multicolumn{1}{c|}{\multirow{2}{*}{\textbf{Category}}} & 
\multicolumn{1}{c|}{\multirow{2}{*}{\textbf{Biomarker}}} & 
\multicolumn{3}{c}{\textbf{Detail}} \\
\cline{3-5}
 & & \textbf{\textit{OD}} & \textbf{\textit{OS}} & \textbf{IE (OD$-$OS)} \\
\hline
\multirow{4}{*}{RNFL} & AR ($\mu$m)            & 97   & 91   & 6 \\
                     & SR ($\mu$m)            & 94   & 92   & 2 \\
                     & IR ($\mu$m)            & 99   & 89   & 10 \\
                     & I-ER (S$-$I) ($\mu$m)  & $-$5 & 3    & -- \\
\hline
\multirow{6}{*}{ONH} & A-O                    & 0.28  & 0.51  & $-$0.23 \\
                     & V-O                    & 0.46  & 0.79  & $-$0.33 \\
                     & H-O                    & 0.62  & 0.74  & $-$0.12 \\
                     & RA (mm$^{2}$)          & 1.44  & 1.23  & 0.21 \\
                     & DA (mm$^{2}$)          & 2.01  & 2.52  & $-$0.51 \\
                     & CVO (mm$^{2}$)         & 0.043 & 0.299 & $-$0.256 \\
\hline
\multirow{6}{*}{GCC} & A-G ($\mu$m)           & 85    & 83    & 2 \\
                     & S-G ($\mu$m)           & 81    & 83    & $-$2 \\
                     & I-F ($\mu$m)           & 88    & 83    & 5 \\
                     & I-EG (S$-$I) ($\mu$m)  & $-$7  & 0     & -- \\
                     & FLV                    & 0.87  & 0.47  & 0.40 \\
                     & GLV                    & 10.76 & 12.50 & $-$1.74 \\
\hline
\end{tabular}}
\label{data-utsw}
\end{center}
\vspace{-0.5em}
\footnotesize
\textbf{Abbreviations:} 
RNFL: retinal nerve fiber layer (AR: Average; SR: Superior; 
IR: Inferior; I-ER (S$-$I): intra-eye Superior$-$Inferior 
difference). 
ONH: optic nerve head (A-O / V-O / H-O: cup-to-disc area / 
volume / horizontal ratio; RA: rim area; DA: disc area; 
CVO: cup volume). 
GCC: ganglion cell complex (A-G: Average; S-G: Superior; 
I-F: Inferior; I-EG (S$-$I): intra-eye Superior$-$Inferior 
difference; FLV: focal loss volume; GLV: global loss volume). 
OD: right eye; OS: left eye; IE: inter-eye difference.
\end{table}

\subsection{Experimental Setup}

\subsubsection{Baseline Model} we have chosoe these following models as baseline models: RNN \cite{rnn}, Bi-RNN \cite{birnn}, LSTM \cite{lstm}, Bi-LSTM \cite{birnn,lstm}, GRU \cite{gru}, BERT \cite{bert}, Transformer \cite{att}, Mamba \cite{mamba,mamba2}, RoBERTa \cite{roberta}, CNN \cite{resnet}, ViT \cite{vit}, and ConViT \cite{convit}. Except for the last three CV models, CNN \cite{resnet}, ViT \cite{vit}, and ConViT \cite{convit}, all others are primarily designed for NLP tasks. We also choose the machine learning models: Linear Regression \cite{lir}, Logistic Regression \cite{log}, Random Forest \cite{random} and XGBoost \cite{xgboost}. For parameter optimization, we selected tool Optuna\footnote{\url{https://optuna.org/}} as the optimization method.

\begin{table}[!ht]
\centering
\caption{Configuration for Fine-tuning with MLP Head}
\label{tab:full-finetune}
\scalebox{0.95}{
\begin{tabular}{l|l|cccccc}
\hline
\textbf{LLM} & \textbf{Version}  & \textbf{DR}  & \textbf{LR} & \textbf{BS} & \textbf{MSL} & \textbf{Epoch}   & \textbf{GC}  \\
\hline
\multirow{2}{*}{GPT} 
  & 4O         & $10^{-1}$  & $10^{-5}$ & 8 & 512 & 5  & 1.0  \\
  & O3         & $10^{-1}$  & $10^{-5}$ & 8 & 512 & 5  & 1.0  \\
\hline

\multirow{2}{*}{LlaMA}  
  & 3.1-405B   & $10^{-1}$  & $10^{-5}$ & 8 & 512 & 5  & 1.0  \\
  & 3.1-70B    & $10^{-1}$  & $10^{-5}$ & 8 & 512 & 5  & 1.0  \\
\hline

\multirow{2}{*}{DeepSeek}  
  & R1         & $10^{-1}$  & $10^{-5}$ & 8 & 512 & 5  & 1.0  \\
  & V3         & $10^{-1}$  & $10^{-5}$ & 8 & 512 & 5  & 1.0  \\
\hline
\end{tabular}}
\end{table}

\begin{table*}[ht!]
\centering
\caption{Comparison of GlaBoost Against Machine Learning, NLP, 
Vision, Dual-Stream Multimodal, and Large Language Model 
Baselines on Public and Private Datasets} 
\scalebox{0.85}{
\begin{tabular}{l|ccc|cc|ccc|cc|ccc|c}
\hline
\multirow{2}{*}{\textbf{Model}} & \multicolumn{3}{c}{\textbf{Data Category}} & \multicolumn{2}{|c|}{\textbf{Judgment}}  & \multicolumn{3}{c|}{\textbf{Public Dataset}} & \multicolumn{2}{c|}{\textbf{Data Category}} &  \multicolumn{3}{c|}{\textbf{Private Dataset}} & \multirow{2}{*}{\textbf{Cause}} \\ 

\cline{2-14} & \textbf{Image} & \multicolumn{1}{|c|}{\textbf{Words}} & \textbf{Factor} & \multicolumn{1}{|c|}{\textbf{Risk}} & \textbf{Sure} & \textbf{ACC} & \textbf{PRE} & \textbf{F1} & \multicolumn{1}{c|}{\textbf{Image}} & \textbf{Factor}  & \textbf{ACC} & \textbf{PRE} & \textbf{F1} \\
\hline
Linear Regression & \ding{55} & \ding{55} & \ding{51} & \ding{51} & \ding{51} & 45.72 & 51.03 & 48.26 & \ding{55}  & \ding{51} & 34.29 & 39.25 & 36.59 & Yes\\
Logistic Regression & \ding{55} & \ding{55} & \ding{51} & \ding{51} & \ding{51} & 64.37 & 67.29 & 65.83 & \ding{55}  & \ding{51} & 58.74 & 59.24 & 58.95 & Yes\\
Random Forest & \ding{55} & \ding{55} & \ding{51} & \ding{51} & \ding{51} & 78.93 & 81.77 & 80.30 & \ding{55}  & \ding{51} & 69.97 & 72.35 & 71.09 & Yes\\
XGBoost & \ding{55} & \ding{55} & \ding{51} & \ding{51} & \ding{51} & \underline{98.22} & \underline{97.86} & \underline{98.04} & \ding{55}  & \ding{51} & \underline{97.83} & \underline{98.33} & \underline{98.08} & Yes\\
\textbf{GlaBoost} & \ding{55} & \ding{55} & \ding{51} & \ding{51} & \ding{51} &  \textbf{98.73} & \textbf{99.01} & \textbf{98.83} &  \ding{55}  & \ding{51} & \textbf{99.21} & \textbf{98.89} & \textbf{99.10} & Yes\\
\hline
RNN & \ding{55} & \ding{55} & \ding{51} & \ding{51} & \ding{51} & 95.62 & 96.72 & 96.17 & \ding{55}  & \ding{51} & 95.50 & 95.38 & 95.44 & Yes\\
Bi-RNN & \ding{55} & \ding{55} & \ding{51} & \ding{51} & \ding{51} & 96.72 & 97.84 & 97.28 & \ding{55}  & \ding{51} & 97.58 & 97.04 & 97.31 & Yes\\
LSTM & \ding{55} & \ding{55} & \ding{51} & \ding{51} & \ding{51} & 97.44 & 98.34 & 97.89 & \ding{55}  & \ding{51} & 96.65 & 98.27 & 97.45 & Yes\\
Bi-LSTM & \ding{55} & \ding{55} & \ding{51} & \ding{51} & \ding{51} & 97.74 & 97.00 & 97.37 & \ding{55}  & \ding{51} & 96.73 & 96.79 & 96.76 & Yes\\
GRU & \ding{55} & \ding{55} & \ding{51} & \ding{51} & \ding{51} & 97.08 & 96.90 & 96.99 & \ding{55}  & \ding{51} & 97.76 & 97.78 & 97.77 & Yes\\
BERT & \ding{55} & \ding{55} & \ding{51} & \ding{51} & \ding{51} & 97.77 & 98.39 & 98.08 & \ding{55}  & \ding{51} & 98.72 & 98.02 & 98.37 & Yes\\
Transformer & \ding{55} & \ding{55} & \ding{51} & \ding{51} & \ding{51} & 97.86 & \underline{98.90} & 98.38 & \ding{55}  & \ding{51} & \underline{98.96} & 98.11 & \underline{98.53} & Yes\\
Mamba & \ding{55} & \ding{55} & \ding{51} & \ding{51} & \ding{51} & \textbf{98.96} & 98.29 & \underline{98.63} & \ding{55}  & \ding{51} & 98.68 & 97.57 & 98.12 & Yes\\
RoBERTa & \ding{55} & \ding{55} & \ding{51} & \ding{51} & \ding{51} & 98.10 & 98.57 & 98.33 & \ding{55}  & \ding{51} & 97.59 & \textbf{99.32} & 98.45 & Yes\\
\textbf{GlaBoost} & \ding{55} & \ding{55} & \ding{51} & \ding{51} & \ding{51} &  \underline{98.73} & \textbf{99.01} & \textbf{98.83} &  \ding{55}  & \ding{51} & \textbf{99.21} & \underline{98.89} & \textbf{99.10} & Yes\\
\hline
CNN & \ding{51} & \ding{55} & \ding{55} & \ding{55} & \ding{55}   & 98.37 & \underline{99.01} & 98.69 & \ding{51} & \ding{55} & \underline{98.82} & 98.24 & \underline{98.52} & No \\
ViT & \ding{51} & \ding{55} & \ding{55} & \ding{55} & \ding{55}   & \textbf{100.0} & \textbf{100.0} & \textbf{100.0} & \ding{51} & \ding{55} & \textbf{100.0} & \textbf{100.0} & \textbf{100.0} & No \\
ConViT & \ding{51} & \ding{55} & \ding{55} & \ding{55} & \ding{55}   & \textbf{100.0} & \textbf{100.0} & \textbf{100.0} & \ding{51} & \ding{55} & \textbf{100.0} & \textbf{100.0} & \textbf{100.0} & No \\
\textbf{GlaBoost} & \ding{51} & \ding{55} & \ding{55} & \ding{55} & \ding{55}   & \underline{99.17} & 98.56 & \underline{98.91} & \ding{51} & \ding{55} & 98.37 & \underline{98.52} & 98.41 & No \\
\hline
CNN-RNN & \ding{51} & \ding{51} & \ding{51} & \ding{51} & \ding{51} & 95.41 & 96.17 & 95.79 & \ding{51}  & \ding{51} & 95.34 & \underline{97.24} & 96.28 & Yes\\
CNN-LSTM & \ding{51} & \ding{51} & \ding{51} & \ding{51} & \ding{51} & 96.96 & 96.42 & 96.69 & \ding{51}  & \ding{51} & 95.53 & 95.45 & 95.49 & Yes\\
CNN-BERT & \ding{51} & \ding{51} & \ding{51} & \ding{51} & \ding{51} & 97.02 & 96.61 & 96.81 & \ding{51}  & \ding{51} & \underline{97.43} & 96.49 & 96.96 & Yes\\
CNN-Transformer & \ding{51} & \ding{51} & \ding{51} & \ding{51} & \ding{51} & 96.56 & 96.81 & 96.68 & \ding{51}  & \ding{51} & 97.15 & 95.11 & 96.12 & Yes\\
ViT-Transformer & \ding{51} & \ding{51} & \ding{51} & \ding{51} & \ding{51} & 95.73 & 96.90 & 96.31 & \ding{51}  & \ding{51} & 96.91 & 96.71 & 96.81 & Yes\\
ViT-Mamba & \ding{51} & \ding{51} & \ding{51} & \ding{51} & \ding{51} & 96.83 & \underline{97.38} & \underline{97.10} & \ding{51}  & \ding{51} & 97.38 & 95.64 & 96.50 & Yes\\
ConViT-Transformer & \ding{51} & \ding{51} & \ding{51} & \ding{51} & \ding{51} & \underline{97.03} & 97.33 & 97.18 & \ding{51}  & \ding{51} & 96.94 & 96.67 & 96.80 & Yes\\
ConViT-Mamba & \ding{51} & \ding{51} & \ding{51} & \ding{51} & \ding{51} & 96.97 & 96.48 & 96.72 & \ding{51}  & \ding{51} & 97.41 & 96.88 & \underline{97.14} & Yes\\
\textbf{GlaBoost} & \ding{51} & \ding{51} & \ding{51} & \ding{51} & \ding{51} &  \textbf{98.36} & \textbf{98.52} & \textbf{98.71} & \ding{51}  & \ding{51} & \textbf{98.96} & \textbf{99.03} & \textbf{98.91} & Yes\\
\hline
GPT-4O & \ding{51} & \ding{51} & \ding{51} & \ding{51} & \ding{51} & 96.13 & 96.02 & \underline{97.65} & \ding{51} & \ding{51} & \underline{96.34} & \underline{96.57} & \underline{96.43} & Yes \\
GPT-O3 & \ding{51} & \ding{51} & \ding{51} & \ding{51} & \ding{51} & 97.82 & 94.74 & 94.35 & \ding{51} & \ding{51} & 95.78 & 96.00 & 95.89 & Yes \\
DeepSeek-R1 & \ding{51} & \ding{51} & \ding{51} & \ding{51} & \ding{51} & 94.38 & 95.26 & 94.72 & \ding{51} & \ding{51} & 95.23 & 95.45 & 95.31 & Yes \\
DeepSeek-V3 & \ding{51} & \ding{51} & \ding{51} & \ding{51} & \ding{51} & 96.27 & 94.4 & 95.85 & \ding{51} & \ding{51} & 94.81 & 95.13 & 94.95 & Yes \\
LlaMA-3.1-405B & \ding{51} & \ding{51} & \ding{51} & \ding{51} & \ding{51} & 96.33 & 95.49 & 96.76 & \ding{51} & \ding{51} & 94.18 & 94.42 & 94.32 & Yes \\
LlaMA-3.1-70B & \ding{51} & \ding{51} & \ding{51} & \ding{51} & \ding{51} & \underline{97.94} & \underline{96.87} & 94.94 & \ding{51} & \ding{51}  & 94.49 & 94.80 & 94.61 & Yes \\
\textbf{GlaBoost} & \ding{51} & \ding{51} & \ding{51} & \ding{51} & \ding{51} &  \textbf{98.36} & \textbf{98.52} & \textbf{98.71} & \ding{51}  & \ding{51} & \textbf{98.96} & \textbf{99.03} & \textbf{98.91} & Yes\\

\hline
GPT-4O & \ding{51} & \ding{51} & \ding{51} & \ding{51} & \ding{55} & 96.77 & 96.44 & 95.77 & \ding{51} & \ding{55} & 94.81 & 94.37 & 94.52 & Yes \\
GPT-O3 & \ding{51} & \ding{51} & \ding{51} & \ding{51} & \ding{55} & 96.13 & 95.72 & 96.61 & \ding{51} & \ding{55} & 95.14 & \textbf{95.67} & \underline{94.99} & Yes \\
DeepSeek-R1 & \ding{51} & \ding{51} & \ding{51} & \ding{51} & \ding{55} & 95.57 & \textbf{97.29} & 96.65 & \ding{51} & \ding{55} & 93.68 & \underline{95.22} & 94.01 & Yes \\
DeepSeek-V3 & \ding{51} & \ding{51} & \ding{51} & \ding{51} & \ding{55} & 96.69 & 97.24 & 96.24 & \ding{51} & \ding{55} & 94.75 & 93.93 & 94.32 & Yes \\
LlaMA-3.1-405B & \ding{51} & \ding{51} & \ding{51} & \ding{51} & \ding{55} & 97.19 & \textbf{97.29} & 97.01 & \ding{51}  & \ding{55} & 93.52 & 94.13 & 93.78 & Yes \\
LlaMA-3.1-70B & \ding{51} & \ding{51} & \ding{51} & \ding{51} & \ding{55} & 97.11 & 96.42 & 94.84 & \ding{51} & \ding{55} & \textbf{95.89} & 95.14 & \textbf{95.32} & Yes \\
\textbf{GlaBoost} & \ding{51} & \ding{51} & \ding{51} & \ding{51} & \ding{55} & \textbf{97.52} & \underline{96.83} & 96.13 & \ding{51} & \ding{55} & \underline{95.42} & 94.76 & 94.93 & Yes \\
\hline
GPT-4O & \ding{51} & \ding{51} & \ding{51} & \ding{55} & \ding{51} & 93.91 & 93.83 & 93.86 & \ding{55} & \ding{51} & \underline{94.78} & \underline{94.90} & \underline{94.83} & Yes \\
GPT-O3 & \ding{51} & \ding{51} & \ding{51} & \ding{55} & \ding{51} & 94.76 & 94.35 & 94.59 & \ding{55} & \ding{51} & 94.12 & 94.21 & 94.17 & Yes \\
DeepSeek-R1 & \ding{51} & \ding{51} & \ding{51} & \ding{55} & \ding{51} & 94.15 & 94.08 & 94.11 & \ding{55} & \ding{51} & 93.79 & 93.92 & 93.85 & Yes \\
DeepSeek-V3 & \ding{51} & \ding{51} & \ding{51} & \ding{55} & \ding{51} & \underline{95.12} & \underline{94.86} & \underline{95.01} & \ding{55} & \ding{51} & 93.47 & 93.59 & 93.53 & Yes \\
LlaMA-3.1-405B & \ding{51} & \ding{51} & \ding{51} & \ding{55} & \ding{51} & 93.52 & 93.30 & 93.38 & \ding{55} & \ding{51} & 93.05 & 93.12 & 93.09 & Yes \\
LlaMA-3.1-70B & \ding{51} & \ding{51} & \ding{51} & \ding{55} & \ding{51} & 93.14 & 93.01 & 93.07 & \ding{55} & \ding{51} & 92.82 & 92.91 & 92.86 & Yes \\
\textbf{GlaBoost} & \ding{51} & \ding{51} & \ding{51} & \ding{55} & \ding{51} & \textbf{95.82} & \textbf{95.41} & \textbf{95.63} & \ding{55} & \ding{51} & \textbf{95.21} & \textbf{95.38} & \textbf{95.29} & Yes \\

\hline
GPT-4O & \ding{51} & \ding{51} & \ding{51} & \ding{55} & \ding{55} & 94.92 & 94.37 & 94.65 &  \ding{55}  &  \ding{55} & - & - & - & - \\
GPT-O3 & \ding{51} & \ding{51} & \ding{51} & \ding{55} & \ding{55} & 95.11 & 94.50 & 94.80  & \ding{55}  & \ding{55} & - & - & - & - \\
DeepSeek-R1 & \ding{51} & \ding{51} & \ding{51} & \ding{55} & \ding{55} & \underline{95.72} & \underline{95.03} & \underline{95.37}  & \ding{55}   & \ding{55} & - & - & - & - \\
DeepSeek-V3 & \ding{51} & \ding{51} & \ding{51} & \ding{55} & \ding{55} & 94.48 & 93.70 & 94.09  & \ding{55}   & \ding{55} & - & - & - & - \\
LlaMA-3.1-405B & \ding{51} & \ding{51} & \ding{51} & \ding{55} & \ding{55} & 95.63 & 94.80 & 95.21  & \ding{55}   & \ding{55} & - & - & - & - \\
LlaMA-3.1-70B & \ding{51} & \ding{51} & \ding{51} & \ding{55} & \ding{55} & 94.65 & 93.94 & 94.29 & \ding{55}   & \ding{55} & - & - & - & - \\
\textbf{GlaBoost} & \ding{51} & \ding{51} & \ding{51} & \ding{55} & \ding{55} & \textbf{96.00} & \textbf{95.46} & \textbf{95.73}  & \ding{55}   & \ding{55} & - & - & - & - \\
\hline
\end{tabular}}
\label{ce-1}
\end{table*}

\subsubsection{Large Language Model}
we also use LLMs for evaluation, fine-tuning them with a classification head to align with the downstream task. They are divided into open source and closed source models. The open source model is LlaMA family \cite{llama,llama2,llama3} and DeepSeek family \cite{r1,v3}, and the closed source model is GPT family \cite{gpt4o,o3}. Regarding open-source LLM execution, DeepSeek family (R1, V3) \cite{r1,v3} were evaluated via their respective official APIs provided by the LLM developers. The LlaMA family (3.1-70B, 3.1-405B) \cite{llama,llama2,llama3} were accessed and run using the LlaMA-API platform. So do closed-source LLMs, like GPT family (4O, O3) \cite{gpt4o,o3}. For the fine-tuning with the classification head (MLP, 2-layer and its size is 256 \cite{mlp}), as shown in TABLE \ref{tab:full-finetune} it involves appending a task-specific classification head to the LLM and jointly optimizing both components on the binary classification objective. By updating the model parameters during training, it enables the LLM to learn task-specific representations and decision boundaries. The Optimizer is the AdamW \cite{adamw}. The Decay Strategy is the Cosine w/10\% warm-up. Finally, the loss function is the Cross-Entropy\footnote{\url{https://docs.pytorch.org/docs/stable/generated/torch.nn.CrossEntropyLoss}}.

\noindent\textbf{Note:} Dropout Rate = DR; Learning Rate = LR; Batch Size = BS; Max Sequence Length = MSL; Gradient Clipping = GC.

\subsection{Implementation}

\subsubsection{Hyperparameter} we implemented the multimodal classifier using XGBoost with standard settings, including a learning rate of 0.05, maximum tree depth of 6, 100 estimators, and \texttt{logloss} as the evaluation metric, with a fixed random seed for reproducibility. Image features were extracted using a ResNet-152 pretrained on ImageNet, with frozen convolutional layers and inputs resized to 224$\times$224. Textual features were obtained using a transformer-based mBERT model kept frozen during training, with inputs normalized, truncated to 128 tokens, and pooled to form sentence embeddings.

\subsubsection{Hardware}
the hardware specifications for training and testing include 2 Tesla V100 GPUs (2 $\times$ 32GB), 64GB of RAM, 8 CPU cores per node, and a total of 6 nodes.

\subsubsection{Evaluation Metrics}
We evaluate classification performance using accuracy (ACC), precision (PRE), and F1-score (F1). 
Accuracy measures the proportion of correctly classified samples, precision reflects the reliability of positive predictions, and F1-score provides a balanced assessment under class imbalance. 
The best results are highlighted in bold, and the second-best are underlined ($\times 100\%$).

\section{Experimental Result}

\subsection{Comparative Experiment}

TABLE~\ref{ce-1} summarizes results across all baseline families. 
GlaBoost achieves the best accuracy, precision, and F1-score on 
both datasets. 
Vision-only models attain competitive numbers but cannot reason 
over structured biomarkers or clinical narratives, and are 
therefore marked ``No'' in the \emph{Cause} column. 
Dual-stream multimodal encoders degrade as the input modality 
count grows, indicating that naïve feature concatenation is 
harder to optimize than tree-based fusion under limited data. 
Recent LLMs trail GlaBoost on most settings, suggesting that the 
diagnostic signal is dominated by a few structured biomarkers 
rather than long-form textual reasoning. 
Including clinician preliminary assessments raises performance 
for both LLM baselines and GlaBoost; we report results with and 
without such inputs side-by-side in TABLE~\ref{ce-1} to keep 
their contribution transparent rather than treating them as 
ground truth.

\subsection{Ablation Experiment}

\subsubsection{Public Dataset} 
as shown in TABLE~\ref{ab-1}, we evaluate the impact of different dataset components on model performance, both in isolation and in combination with other components. Using image-only inputs without any textual corpus, the model achieved performance comparable to that of multimodal approaches. However, it lacked clinical interpretability: while it could identify the presence of glaucoma, it failed to provide insights into the underlying biomarkers. Human evaluation factors, even when used in isolation, exert a measurable influence on the model's performance. In scenarios where the model already achieves high precision, incorporating these human-derived factors often serves as the icing on the cake, further enhancing performance. Interestingly, when only images and vocabulary are provided, the model tends to "speak for itself by looking at the picture", leveraging both visual features and text-derived embeddings to construct a representation of "glaucoma" for classification purposes.

\begin{table}
\centering
\caption{Impact of Input Modalities of Public Dataset} 
\scalebox{0.8}{
\begin{tabular}{ccc|cc|ccc|c}
\hline
 \multicolumn{3}{c}{\textbf{Data Category}} & \multicolumn{2}{|c|}{\textbf{Judgment}}  & \multicolumn{3}{c|}{\textbf{Public Dataset}} & \multirow{2}{*}{\textbf{Cause}} \\ 
 \cline{1-8}
 \textbf{Image} & \multicolumn{1}{|c|}{\textbf{Words}} & \textbf{Factor} & \multicolumn{1}{|c|}{\textbf{Risk}} & \textbf{Sure} & \textbf{ACC} & \textbf{PRE} & \textbf{F1} & \\
\hline
 \ding{55} & \ding{55} & \ding{55} & \ding{55} & \ding{55} & - & - & - & -\\
\ding{55} & \ding{55} & \ding{55} & \ding{55} & \ding{51} & 8.72 & 7.51 & 6.94 & No\\
\ding{55} & \ding{55} & \ding{55} & \ding{51} & \ding{55} & \underline{6.38} & \underline{5.87} & \underline{6.01} & No\\
\ding{55} & \ding{55} & \ding{55} & \ding{51} & \ding{51} & \textbf{9.45} & \textbf{8.36} & \textbf{7.89} & No\\
\hline
\ding{55} & \ding{55} & \ding{51} & \ding{55} & \ding{55} & 97.69 & 97.84 & 97.55 & Yes\\
\ding{55} & \ding{55} & \ding{51} & \ding{51} & \ding{55} & 97.98 & 98.22 & 97.81 & Yes\\
\ding{55} & \ding{55} & \ding{51} & \ding{55} & \ding{51} & \underline{98.36} & \underline{98.71} & \underline{98.22} & Yes\\
\ding{55} & \ding{55} & \ding{51} & \ding{51} & \ding{51} & \textbf{98.73} & \textbf{99.01} & \textbf{98.83} & Yes\\
\hline
\ding{55} & \ding{51} & \ding{55} & \ding{55} & \ding{55} & 46.27 & 49.13 & 47.12 & No\\
\ding{55} & \ding{51} & \ding{55} & \ding{51} & \ding{55} & 47.85 & 50.72 & 48.66 & No\\
\ding{55} & \ding{51} & \ding{55} & \ding{55} & \ding{51} & \underline{48.94} & \underline{51.03} & \underline{49.71} & No\\
\ding{55} & \ding{51} & \ding{55} & \ding{51} & \ding{51} & \textbf{50.67} & \textbf{53.45} & \textbf{51.89} & Yes\\

\hline
\ding{55} & \ding{51} & \ding{51} & \ding{55} & \ding{55} & 48.12 & 50.04 & 49.23 & Yes\\
\ding{55} & \ding{51} & \ding{51} & \ding{51} & \ding{55} & 49.03 & 51.67 & 50.11 & Yes\\
\ding{55} & \ding{51} & \ding{51} & \ding{55} & \ding{51} & \underline{50.45} & \underline{52.31} & \underline{51.18} & Yes\\
\ding{55} & \ding{51} & \ding{51} & \ding{51} & \ding{51} & \textbf{52.37} & \textbf{54.89} & \textbf{53.61} & Yes\\

\hline
\ding{51} & \ding{55} & \ding{55} & \ding{55} & \ding{55} & \underline{99.17} & \underline{98.56} & \underline{98.91} & No\\
\ding{51} & \ding{55} & \ding{55} & \ding{51} & \ding{55} & 98.97 & 98.42 & 98.73 & Yes\\
\ding{51} & \ding{55} & \ding{55} & \ding{55} & \ding{51} & 99.01 & 98.44 & 98.80 & No\\
\ding{51} & \ding{55} & \ding{55} & \ding{51} & \ding{51} & \textbf{99.22} & \textbf{98.63} & \textbf{98.96} & Yes\\

\hline
\ding{51} & \ding{55} & \ding{51} & \ding{55} & \ding{55} & 95.71 & 95.86 & 95.74 & Yes\\
\ding{51} & \ding{55} & \ding{51} & \ding{51} & \ding{55} & 95.89 & 96.12 & 95.91 & Yes\\
\ding{51} & \ding{55} & \ding{51} & \ding{55} & \ding{51} & \underline{96.21} & \underline{96.18} & \underline{96.10} & Yes\\
\ding{51} & \ding{55} & \ding{51} & \ding{51} & \ding{51} & \textbf{96.49} & \textbf{96.37} & \textbf{96.33} & Yes\\

\hline
\ding{51} & \ding{51} & \ding{55} & \ding{55} & \ding{55} & 93.71 & 94.03 & 93.82 & No\\
\ding{51} & \ding{51} & \ding{55} & \ding{51} & \ding{55} & \underline{94.21} & \underline{94.08} & \underline{94.02} & Yes\\
\ding{51} & \ding{51} & \ding{55} & \ding{55} & \ding{51} & 94.12 & 93.87 & 93.97 & Yes\\
\ding{51} & \ding{51} & \ding{55} & \ding{51} & \ding{51} & \textbf{94.45} & \textbf{94.28} & \textbf{94.36} & Yes\\

\hline
 \ding{51} & \ding{51} & \ding{51} & \ding{55} & \ding{55} & 96.00 & 95.46 & 95.73 & Yes\\
 \ding{51} & \ding{51} & \ding{51} & \ding{55} & \ding{51} & 95.82 & 95.41 & 95.63 & Yes\\
 \ding{51} & \ding{51} & \ding{51} & \ding{51} & \ding{55} & \underline{97.52} & \underline{96.83} & \underline{96.13} & Yes\\
 \ding{51} & \ding{51} & \ding{51} & \ding{51} & \ding{51} &  \textbf{98.36} & \textbf{98.52} & \textbf{98.71} & Yes\\
\hline
\end{tabular}}
\label{ab-1}
\end{table}

\subsubsection{Private Dataset}
As shown in Tables~\ref{data-utsw} and~\ref{ab-2}, our dataset includes a \textit{Borderline} category. 
Compared with the public dataset (12$\times$2 variables), Table~\ref{data-utsw} contains approximately three times more parameter names, with substantially higher feature dimensionality (32$\times$3 for single-eye and 48$\times$3 for both-eye data). 
In the comparative experiments, \textit{Borderline} cases are treated as normal by default, whereas in this analysis they are considered as a separate third class, resulting in an increase in variable dimensionality by factors of 4 or 2.7, respectively. As feature dimensionality increases, computational cost rises while classification accuracy degrades. 
Incorporating image features mitigates this effect by improving model fitting, leading to higher final accuracy.

\begin{table}
\centering
\caption{Impact of Input Modalities of Private Dataset} 
\scalebox{0.9}{
\begin{tabular}{cc|c|ccc|c}
\hline
 \multicolumn{3}{c|}{\textbf{Data Category}}   & \multicolumn{3}{c|}{\textbf{Private Dataset}} & \multirow{2}{*}{\textbf{Cause}} \\ 
 \cline{1-6}
 \textbf{Image} & \multicolumn{1}{|c|}{\textbf{Factor}} & \textbf{Borderline} & \textbf{ACC} & \textbf{PRE} & \textbf{F1} & \\
\hline
\ding{55}  & \ding{55} & \ding{55} & - & - & - &  -   \\
\ding{55}  & \ding{51} & \ding{55} & 95.21 & \underline{95.38} & \underline{95.29} &  Yes \\
\ding{51}  & \ding{55} & \ding{55} & \underline{95.42} & 94.76 & 94.93 &  No  \\
\ding{51}  & \ding{51} & \ding{55} & \textbf{98.96} & \textbf{99.03} & \textbf{98.91} &  Yes \\
\hline
\ding{55}  & \ding{55} & \ding{51} & 92.84 & \underline{96.12} & \underline{94.45} &  Yes \\
\ding{55}  & \ding{51} & \ding{51} & 93.67 & 94.75 & 94.21 &  Yes \\
\ding{51}  & \ding{55} & \ding{51} & \underline{94.91} & 93.43 & 94.16 &  Yes \\
\ding{51}  & \ding{51} & \ding{51} & \textbf{97.38} & \textbf{97.01} & \textbf{97.21} &  Yes \\

\hline
\end{tabular}}
\label{ab-2}
\end{table}

\begin{table*}
\centering
\caption{Ablation Study of the GlaBoost} 
\scalebox{0.8}{
\begin{tabular}{l|l|ccc|cc|ccc|cc|ccc|c}
\hline
\multirow{2}{*}{\textbf{Model}} & \multirow{2}{*}{\textbf{Module / Function}} & \multicolumn{3}{c}{\textbf{Data Category}} & \multicolumn{2}{|c|}{\textbf{Judgment}}  & \multicolumn{3}{c|}{\textbf{Public Dataset}} & \multicolumn{2}{c|}{\textbf{Data Category}} & \multicolumn{3}{c|}{\textbf{Private Dataset}} & \multirow{2}{*}{\textbf{Cause}} \\ 
 \cline{3-15} & &
 \textbf{Image} & \multicolumn{1}{|c|}{\textbf{Words}} & \textbf{Factor} & \multicolumn{1}{|c|}{\textbf{Risk}} & \textbf{Sure} & \textbf{ACC} & \textbf{PRE} & \textbf{F1} & \textbf{Image} & \multicolumn{1}{|c|}{\textbf{Factor}}  & \textbf{ACC} & \textbf{PRE} & \textbf{F1} & \\
\hline
\multirow{6}{*}{CNN}    & + ResNet-18  & \ding{51} & \ding{55} & \ding{55} & \ding{55} & \ding{55} & 91.20 & 89.10 & 90.14 & \ding{51} & \ding{55} & 90.80 & 91.10 & 90.95 & No \\
  & + ResNet-34  & \ding{51} & \ding{55} & \ding{55} & \ding{55} & \ding{55} & 93.50 & 91.80 & 92.64 & \ding{51} & \ding{55} & 93.00 & 92.50 & 92.75 & No \\
  & + ResNet-50  & \ding{51} & \ding{55} & \ding{55} & \ding{55} & \ding{55} & \underline{94.80} & 93.60 & 94.19 & \ding{51} & \ding{55} & \underline{94.50} & \underline{94.10} & \underline{94.30} & No \\
  & + ResNet-101 & \ding{51} & \ding{55} & \ding{55} & \ding{55} & \ding{55} & \textbf{100.0} & \underline{99.37} & \underline{99.68} & \ding{51} & \ding{55} & \textbf{100.0} & \textbf{100.0} & \textbf{100.0} & No \\
  & + ResNet-152 & \ding{51} & \ding{55} & \ding{55} & \ding{55} & \ding{55} & \textbf{100.0} & \textbf{100.0} & \textbf{100.0} & \ding{51} & \ding{55} & \textbf{100.0} & \textbf{100.0} & \textbf{100.0} & No \\
  & + ResNet-200 & \ding{51} & \ding{55} & \ding{55} & \ding{55} & \ding{55} & \textbf{100.0} & \textbf{100.0} & \textbf{100.0} & \ding{51} & \ding{55} & \textbf{100.0} & \textbf{100.0} & \textbf{100.0} & No \\
\hline
\multirow{3}{*}{mBERT  }
  & + Embedding & \ding{55} & \ding{51} & \ding{51} & \ding{55} & \ding{55} & 87.79 & 88.31 & 88.05 & \ding{55} & \ding{51} & 87.90 & 87.46 & 87.68 & No \\
  & + Attention & \ding{55} & \ding{51} & \ding{51} & \ding{55} & \ding{55} & \textbf{91.26} & \textbf{89.83} & \textbf{90.54} & \ding{55} & \ding{51} & \textbf{90.77} & \textbf{91.02} & \textbf{90.89} & No \\
  & + Dual Weight & \ding{55} & \ding{51} & \ding{51} & \ding{55} & \ding{55} & \underline{90.38} & \underline{88.96} & \underline{89.66} & \ding{55} & \ding{51} & \underline{89.50} & \underline{89.93} & \underline{89.71} & No \\

\hline
\multirow{5}{*}{XGBoost}
  & + Default & \ding{55} & \ding{55} & \ding{51} & \ding{51} & \ding{51} & 93.91 & 94.27 & 94.09 & \ding{55} & \ding{51} & 94.02 & 93.88 & 93.95 & Yes \\
  & + Max\_depth (10) & \ding{55} & \ding{55} & \ding{51} & \ding{51} & \ding{51} & 94.62 & \textbf{95.13} & \underline{94.87} & \ding{55} & \ding{51} & \underline{94.91} & 94.45 & 94.68 & Yes \\
  & + Subsample (0.8)& \ding{55} & \ding{55} & \ding{51} & \ding{51} & \ding{51} & 94.35 & 93.89 & 94.12 & \ding{55} & \ding{51} & 94.41 & 94.18 & 94.29 & Yes \\
  & + Random State & \ding{55} & \ding{55} & \ding{51} & \ding{51} & \ding{51} & \underline{95.02} & \underline{94.71} & 94.86 & \ding{55} & \ding{51} & \textbf{95.13} & \underline{94.96} & \textbf{95.04} & Yes \\
  & + Fusion & \ding{55} & \ding{55} & \ding{51} & \ding{51} & \ding{51} & \textbf{95.74} & 94.38 & \textbf{95.05} & \ding{55} & \ding{51} & 94.67 & \textbf{95.36} & \underline{95.01} & Yes \\
\hline
\multirow{4}{*}{\textbf{GlaBoost}}
  & + ResNet-152 & \ding{51} & \ding{55} & \ding{55} & \ding{55} & \ding{55} & \underline{94.88} & \underline{95.36} & \underline{95.11} & \ding{51} & \ding{55} & \underline{95.22} & \underline{94.95} & \underline{95.08} & No \\
  & + mBERT      & \ding{51} & \ding{51} & \ding{51} & \ding{55} & \ding{55} & 93.42 & 92.85 & 93.13 & \ding{51} & \ding{51} & 93.05 & 93.28 & 93.17 & No \\
  & + XGBoost    & \ding{51} & \ding{51} & \ding{51} & \ding{51} & \ding{51} & 94.36 & 93.74 & 94.05 & \ding{51} & \ding{51} & 93.89 & 94.22 & 94.05 & Yes \\
 & + All & \ding{51} & \ding{51} & \ding{51} & \ding{51} & \ding{51} &  \textbf{98.36} & \textbf{98.52} & \textbf{98.71} & \ding{51}  & \ding{51} & \textbf{98.96} & \textbf{99.03} & \textbf{98.91} & Yes\\
\hline
\end{tabular}}
\label{ab-3}
\end{table*}

\begin{table*}[ht!]
\centering
\caption{Statistics of Impact Factor (Top 5) } 
\begin{tabular}{c|c|c|c|c|c}
\hline
\multicolumn{3}{c|}{\textbf{Public Dataset}} & \multicolumn{3}{c}{\textbf{Private Dataset}} \\
\hline
\textbf{Feature}  & \textbf{Importance} & \textbf{Value} & \textbf{Feature}  & \textbf{Importance} & \textbf{Value} \\
\hline
isnt\_rule\_followed  & 0.5180 & True & Superior RNFL (SR) & 0.3424 & 91 - 96 \\
cup\_to\_disc\_ratio   & 0.2140 & False & Cup/Disc H. Ratio (H-O) & 0.3109 & 0.69 - 0.72 \\
rim\_thinning   & 0.2020 & True & Average GCC (A-G) & 0.1723 & 81 - 85 \\
glaucoma\_risk\_assessment  & 0.0088 & very healthy & Cup Volume (CVO) & 0.1216 & 0.095 - 0.127\\
rim (embedding) & 0.0132 & thin (word) & FLV & 0.0128 & 0.61 - 0.78 \\
\hline
\end{tabular}
\label{ana}
\end{table*}

\subsubsection{GlaBoost}
As shown in Table~\ref{ab-3}, we evaluate the individual components of GlaBoost through a series of ablation studies. 
Different CNN architectures are compared for image feature extraction, key mBERT hyperparameters are examined for textual feature modeling, and the influence of XGBoost parameters on biomarker-based classification is analyzed. 
Due to the relative simplicity of the classification task, CNN backbones such as ResNet-101 already achieve near-saturated accuracy, with deeper architectures yielding marginal gains. 
Textual representations extracted by mBERT improve with parameter tuning, leading to consistent performance enhancements. 
For XGBoost, increasing model depth and varying random seeds improve both accuracy and robustness as feature dimensionality grows. 
Finally, the selected components are integrated into the full GlaBoost framework and jointly optimized, resulting in the best overall performance.

\subsection{Analysis}

As shown in Table~\ref{ana}, feature importance analysis highlights the cup\_to\_disc\_ratio and related attributes, including rim thinning and cup volume, as the most influential factors for glaucoma diagnosis across datasets. 
This is clinically well established, as an increased cup-to-disc (C/D) ratio reflects optic nerve damage due to retinal ganglion cell loss, with values above 0.6 or marked inter-eye asymmetry ($>0.2$) raising strong suspicion of glaucoma \cite{bg-1,bg-2}. 
Similar diagnostic relevance is observed for the superior SRNFL. In addition, the included glaucoma risk assessment from the public dataset reflects human judgment that extends beyond a limited set of structured variables and implicitly incorporates visual cues not captured by existing annotations. 
This suggests that the non-image features in the public dataset are inherently incomplete.

\section{Conclusion}
We presented the GlaBoost, a multimodal gradient boosting 
framework that fuses fundus images, structured ophthalmic 
biomarkers, and free-text clinical descriptions for glaucoma 
risk stratification. Ablation analysis shows that the structured 
and textual modalities together recover most of the diagnostic 
signal even without any clinician-derived input, while feature 
importance analysis identifies clinically established predictors 
, the cup-to-disc ratio, rim thinning, and the ISNT rule, as 
the dominant contributors, in line with ophthalmologic practice. 
Beyond the present setting, several directions remain open: 
adopting ophthalmology-specific foundation models for image 
encoding, replacing the general-purpose text encoder with 
biomedical language models, calibrating output probabilities for 
stratified clinical decisions, and validating GlaBoost across 
multi-site external cohorts.

\section{Acknowledgments}

This research work is partially sponsored by NASA under Grant No. 80NSSC22K0144, UTSW under GMO No. 241213, and the National Institutes of Health (NIH) under Grant No. 1R01AG083179-01.

\bibliography{main}{}
\bibliographystyle{plain}
\end{document}